\documentclass{article} 
\usepackage{iclr2025_conference,times}

\usepackage{amsmath,amsfonts,bm}

\def\eqref#1{equation~\ref{#1}}

\def\1{\bm{1}}

\DeclareMathAlphabet{\mathsfit}{\encodingdefault}{\sfdefault}{m}{sl}
\SetMathAlphabet{\mathsfit}{bold}{\encodingdefault}{\sfdefault}{bx}{n}

\usepackage{hyperref}
\usepackage{url}
\usepackage{twemojis}

\usepackage{tabularx}
\usepackage{adjustbox}
\usepackage{booktabs}
\usepackage{multirow}
\usepackage[separate-uncertainty]{siunitx}
\newrobustcmd\B{\DeclareFontSeriesDefault[rm]{bf}{b}\bfseries}
\usepackage[tableposition=top]{caption}

\usepackage{graphicx}
\usepackage{subcaption}

\title{Personalized Federated Fine-Tuning of Vision Foundation Models for Healthcare}

\author{Adam Tupper \\
Institut intelligence et donn\'ees (IID) \\ Universit\'e Laval \\ Mila - Quebec AI Institute \\
\texttt{adam.tupper.1@ulaval.ca} \\
\And
Christian Gagn\'e \\
Institut intelligence et donn\'ees (IID) \\ Universit\'e Laval \\ Mila - Quebec AI Institute \\
\texttt{christian.gagne@gel.ulaval.ca} \\
}

\iclrfinalcopy 
\begin{document}

\maketitle

\begin{abstract}
Foundation models open up new possibilities for the use of AI in healthcare.  However, even when pre-trained on health data, they still need to be fine-tuned for specific downstream tasks. Furthermore, although foundation models reduce the amount of training data required to achieve good performance, obtaining sufficient data is still a challenge. This is due, in part, to restrictions on sharing and aggregating data from different sources to protect patients' privacy. One possible solution to this is to fine-tune foundation models via federated learning across multiple participating clients (i.e., hospitals, clinics, etc.). In this work, we propose a new personalized federated fine-tuning method that learns orthogonal LoRA adapters to disentangle general and client-specific knowledge, enabling each client to fully exploit both their own data and the data of others. Our preliminary results on real-world federated medical imaging tasks demonstrate that our approach is competitive against current federated fine-tuning methods.
\end{abstract}

\section{Introduction}

Recent breakthroughs in deep learning—driven by scaling model architectures and training data—have significantly advanced performance across domains, including medical imaging. However, we often assume that this data can be aggregated so that training is carried out in one central location, which is infeasible in healthcare settings due to strict privacy regulations. Federated learning \citep{mcmahan2017} has emerged as an effective technique to enable model training under such constraints. It enables multiple institutions, such as hospitals, to collaboratively train a shared model without exchanging sensitive data.

While federated learning addresses data-sharing constraints, another promising direction is the use of vision foundation models. These models leverage large-scale pre-training to improve performance on downstream tasks and generalize well from limited labeled data. This is an advantage in  medical imaging, where annotated datasets are scarce and costly to obtain. However, fine-tuning vision foundation models via federated learning introduces new challenges: their size increases communication overhead and the compute requirements for the clients, and they do not inherently resolve data heterogeneity across clients.

To address these limitations, recent work has explored combining Low-Rank Adaptation (LoRA) \citep{hu2022b} with federated learning and partial model personalization, where only a subset of model parameters is shared and updated globally \citep{qi2024b, sun2023c, yang2024c, zhang2024f, guo2024, mitra2025}. This approach reduces training costs and helps mitigate data heterogeneity. Building on these ideas, we propose \textbf{Fed}erated \textbf{O}rthogonal \textbf{P}ersonalized \textbf{A}dapter \textbf{L}earning (FedOPAL), a novel method that extends personalized federated fine-tuning by learning orthogonal LoRA adapters to disentangle client-independent and client-specific knowledge. This enables each client to fully exploit both their own data and the data of others. We demonstrate the potential of our approach on real-world federated medical imaging datasets, showing competitive performance compared to existing methods.

\section{Personalized Orthogonal Adapters}

\begin{figure}[t]
    \centering
    \includegraphics[width=\linewidth]{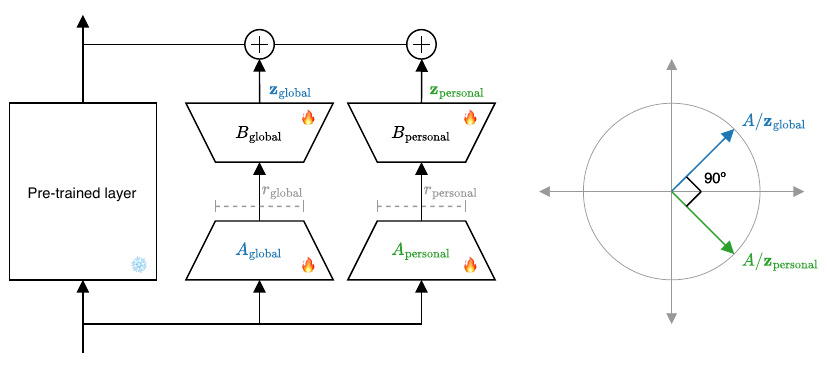}
    \caption{An overview of FedOPAL. Left: Two trainable LoRA adapters (\texttwemoji{fire}) are added to selected layers of a frozen pre-trained model (\texttwemoji{snowflake}): a global adapter, collaboratively updated by all clients, and a personal adapter, which remains private to each client. Right: Orthogonality is enforced between \textit{either} the LoRA $A$ matrix weights \textit{or} the representations learned by each adapter to promote disentangled feature learning.}
    \label{fig:fedopal_diagram}
\end{figure}

We propose a dual orthogonal adapter approach to personalized federated fine-tuning, illustrated in Fig. \ref{fig:fedopal_diagram}. Similar to FedDPA \citep{yang2024c}, all clients participate in training a shared global adapter whilst also training their own personal adapters. However, unlike FedDPA, where the personal adapters are designed to the personalize the global model to the clients' data, we train the adapters simultaneously and aim to encourage the two adapters to learn separate features sets. The global adapter should learn general, client-independent features and the personal adapters client-specific features. To enforce this separation, we propose two methods: FedOPAL-W, which applies orthogonal regularization to the adapter weights, and FedOPAL-R, which applies it to the learned representations.

Using a single LoRA adapter, for a pre-trained weight matrix $W_0 \in \mathbb{R}^{d \times k}$, the weights are updated with a low-rank decomposition $W_0 + \Delta W = W_0 + BA$, where $B \in \mathbb{R}^{d \times r}$, $A \in \mathbb{R}^{r \times k}$ and the rank $r \ll \min(d, k)$. During training, $W_0$ is frozen, while $A$ and $B$  are updated. With two adapters, one global and one personal,  $\Delta W = W_0 + B_\text{global}A_\text{global} + B_\text{personal}A_\text{personal}$. During each round of local client training, both adapters are enabled and trained simultaneously. At the end of the round, only the parameters of the global adapter are updated and aggregated using FedAvg \citep{mcmahan2017}.

The orthogonal weight regularization is motivated by O-LoRA, a method for continual learning using incremental LoRA adapters \citep{wang2023f}. They introduce orthogonal regularization to mitigate catastrophic forgetting by constraining the gradient updates of the current task to be orthogonal to the gradient subspace of past tasks. Here, we apply the same methodology, but to mitigate the interference of the personal adapters on the global adapter. The orthogonal loss over the $N$ layers of the model being fine-tuned is given by

\begin{equation}
    \ell_\text{orth} = \frac{1}{N} \sum_{i=1}^N \left| (A_\text{global}^i)^T A_\text{personal}^i \right|.
\end{equation}

Orthogonal weight regularization is designed to mitigate interference between the global and local adapters and indirectly encourage them to learn different things. However, it is still possible that they encode similar features despite the constraint on their weights. To directly target the features learned by the adapters, we propose an alternative approach that acts directly on the representations they learn. Specifically, we add a regularization term that acts on the representations $\mathbf{z}_\text{global}$ and $\mathbf{z}_\text{personal}$ learned by the adapters for each of the $B$ examples in the mini-batch. This is achieved by minimizing the mean absolute cosine similarity between each pair of adapters over the $N$ layers of the model being fine-tuned:

\begin{equation}
    \ell_\text{orth} = \frac{1}{NB}\sum_{i=1}^N\sum_{j=1}^B \left| \cos(\mathbf{z}_\text{global}^{i,j}, \mathbf{z}_\text{personal}^{i,j}) \right|.
\end{equation}

The total training loss is given by $\ell_\text{total} = \ell_\text{task} + \lambda \ell_\text{orth}$, where $\lambda$ is a tunable scaling factor and $\ell_\text{task}$ is the cross entropy loss.

\section{Experiments \& Preliminary Results}

We performed evaluations on two challenging real-world federated tasks: dermatoscopic image classification using the six-client Fed-ISIC 2019 dataset \citep{ogierduterrail2022} and patch-level classification of breast cancer metastases in whole-slide images of histological lymph node sections using the five-client Camelyon17-WILDS dataset \citep{koh2021}.\footnote{The Fed-ISIC and Camelyon17-WILDS datasets are derived from the ISIC 2019 Challenge \citep{codella2018, tschandl2018, hernandez-perez2024} and Camelyon17 \citep{bandi2019} datasets, respectively.} In addition to feature shift, the Fed-ISIC 2019 dataset has significant label shift and uneven sample sizes between clients, as shown in Fig. 2. While the Camelyon17-WILDS dataset provides a large number of image patches per client (Fig. 2), sample size and label shift are still problematic because the data is sourced from only 7--10 patients. More details about each dataset, including examples of feature shift, are provided in Appendix \ref{app:datasets}.

For each task, we fine-tuned ViT-Tiny models with 9.7 million parameters that were pre-trained on the ImageNet-21K dataset by adding LoRA adapters to the query, key and value projection matrices in each transformer block. We compared our proposed methods against local (a single client's data) and centralized (all clients' data) training, training without regularization (FedPAL), and state-of-the-art LoRA-based federated learning methods. For both tasks, we measured performance using balanced accuracy to account for class imbalance. Additional information about our experimental setup and the baselines are included in Appendix \ref{app:experimental-setup}.

\begin{figure}[t]
    \centering
    \begin{subfigure}[b]{0.49\textwidth}
        \centering
        \includegraphics[width=\linewidth]{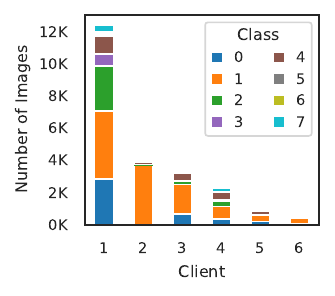}
    \end{subfigure}
    \hfill 
    \begin{subfigure}[b]{0.49\textwidth}
        \centering
        \includegraphics[width=\linewidth]{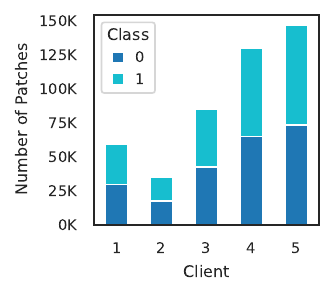}
    \end{subfigure}
    \caption{The client class distributions for the ISIC 2019 (left) and Camelyon17 (right) datasets.}
    \label{fig:dataset_class_distributions}
\end{figure}

Our preliminary results for the Fed-ISIC 2019 and Camelyon17-WILDS tasks are presented in tables \ref{tab:isic-results} and \ref{tab:camelyon17-results}, respectively. While our methods are competitive compared to the other methods, no one method is consistently the best across all clients on either task. Further investigation is needed to understand this, but it indicates room for additional improvement.

\begin{table}[t]
\centering
\caption{The balanced client test accuracies (mean and std. dev. over three runs) for each method on the Fed-ISIC 2019 task.}
\label{tab:isic-results}
\begin{adjustbox}{width=\textwidth}
\begin{tabular}{
    @{}
    l
    S[table-format = 1.3(3), table-align-uncertainty = false]
    S[table-format = 1.3(3), table-align-uncertainty = false]
    S[table-format = 1.3(3), table-align-uncertainty = false]
    S[table-format = 1.3(3), table-align-uncertainty = false]
    S[table-format = 1.3(3), table-align-uncertainty = false]
    S[table-format = 1.3(3), table-align-uncertainty = false]
    S[table-format = 1.3(3), table-align-uncertainty = false]
    c
    @{}
}
\toprule
\multicolumn{1}{c}{\multirow{2}{*}{}} & \multicolumn{6}{c}{Client} \\ \cmidrule(lr){2-7} 
\multicolumn{1}{c}{Method} & \multicolumn{1}{c}{1} & \multicolumn{1}{c}{2} & \multicolumn{1}{c}{3} & \multicolumn{1}{c}{4} & \multicolumn{1}{c}{5} & \multicolumn{1}{c}{6} & \multicolumn{1}{c}{Avg.} & Avg. Rank \\
\midrule
Centralized & 0.758 (0.007)   & 0.665 (0.052)   & 0.712 (0.103)   & 0.615 (0.071)   & 0.647 (0.046)   & 0.787 (0.044)   & 0.697 (0.022)   & n/a    \\
Local       & \B0.772 (0.008) & 0.850 (0.032)   & 0.728 (0.086)   & 0.614 (0.069)   & 0.653 (0.006)   & 0.356 (0.127)   & 0.656 (0.175)   & 5.00   \\
\cmidrule{1-8}
FedIT       & 0.755 (0.010)   & 0.775 (0.040)   & 0.731 (0.092)   & 0.514 (0.007)   & 0.474 (0.031)   & 0.685 (0.030)   & 0.656 (0.127)   & 6.17   \\
FedSA       & 0.753 (0.020)   & 0.872 (0.010)   & 0.753 (0.090)   & 0.647 (0.029)   & 0.665 (0.020)   & 0.732 (0.075)   & \B0.737 (0.087) & 3.67   \\
FFA-LoRA    & 0.731 (0.010)   & 0.797 (0.030)   & 0.758 (0.017)   & 0.482 (0.038)   & 0.485 (0.032)   & \B0.770 (0.040) & 0.670 (0.140)   & 5.50   \\
FedDPA      &  0.755 (0.010)   & 0.824 (0.059)   & 0.758 (0.076)   & 0.578 (0.010)   & 0.644 (0.025)   & 0.753 (0.026)   & 0.719 (0.092)  & 4.33   \\
\cmidrule{1-8}
FedPAL      & 0.770 (0.025)   & \B0.931 (0.040) & 0.713 (0.076)   & \B0.657 (0.015) & 0.674 (0.021)   & 0.614 (0.056)   & 0.726 (0.113)   & 3.50   \\
FedOPAL-W   & 0.744 (0.013)   & 0.865 (0.075)   & 0.758 (0.093)   & 0.624 (0.017)   & \B0.687 (0.013) & 0.619 (0.160)   & 0.716 (0.111)   & 3.83   \\
FedOPAL-R   & 0.761 (0.005)   & 0.883 (0.053)   & \B0.814 (0.020) & 0.632 (0.042)   & 0.653 (0.054)   & 0.615 (0.192)   & 0.726 (0.126)   & \B3.17 \\
\bottomrule
\end{tabular}
\end{adjustbox}
\end{table}

\begin{table}[t]
\centering
\caption{The balanced client test accuracies (mean and std. dev. over three runs) for each method on the Camelyon17-WILDS task.}
\label{tab:camelyon17-results}
\begin{adjustbox}{width=\textwidth}
\begin{tabular}{
    @{}
    l
    S[table-format = 1.3(3), table-align-uncertainty = false]
    S[table-format = 1.3(3), table-align-uncertainty = false]
    S[table-format = 1.3(3), table-align-uncertainty = false]
    S[table-format = 1.3(3), table-align-uncertainty = false]
    S[table-format = 1.3(3), table-align-uncertainty = false]
    S[table-format = 1.3(3), table-align-uncertainty = false]
    c
    @{}
}
\toprule
\multicolumn{1}{c}{\multirow{2}{*}{}} & \multicolumn{5}{c}{Client} \\ \cmidrule(lr){2-6} 
\multicolumn{1}{c}{Method} & \multicolumn{1}{c}{1} & \multicolumn{1}{c}{2} & \multicolumn{1}{c}{3} & \multicolumn{1}{c}{4} & \multicolumn{1}{c}{5} & \multicolumn{1}{c}{Avg.} & Avg. Rank \\
\midrule
Centralized & 0.867 (0.051)   & 0.874 (0.042)   & 0.923 (0.026)   & 0.972 (0.004)   & 0.892 (0.039)   & 0.905 (0.020) & n/a  \\
Local       & 0.826 (0.033)   & 0.800 (0.042)   & 0.890 (0.011)   & 0.929 (0.024)   & 0.778 (0.054)   & 0.844 (0.066) & 6.00 \\
\cmidrule{1-7}
FedIT       & \B0.918 (0.017) & 0.895 (0.022)   & 0.901 (0.008)   & 0.924 (0.020)   & 0.772 (0.101)   & 0.882 (0.071) & 4.20 \\
FedSA       & 0.750 (0.028)   & 0.828 (0.035)   & \B0.916 (0.011) & 0.940 (0.027)   & 0.788 (0.050)   & 0.844 (0.081) & 4.20 \\
FFA-LoRA    & 0.893 (0.003)   & 0.870 (0.053)   & 0.901 (0.012)   & 0.921 (0.021)   & 0.771 (0.096)   & 0.871 (0.069) & 5.20 \\
FedDPA      & 0.899 (0.023)   & \B0.931 (0.010) & 0.914 (0.010)   & 0.939 (0.017)   & 0.859 (0.056) & \B0.909 (0.038) & \B2.40 \\
\cmidrule{1-7}
FedPAL      & 0.768 (0.064)   & 0.822 (0.059)   & 0.907 (0.018)   & \B0.950 (0.005) & 0.753 (0.110)   & 0.840 (0.096) & 5.00 \\
FedOPAL-W   & 0.791 (0.014)   & 0.824 (0.065)   & 0.886 (0.029)   & 0.946 (0.011)   & 0.802 (0.070)   & 0.850 (0.071) & 5.00 \\
FedOPAL-R   & 0.824 (0.017)   & 0.808 (0.028)   & 0.902 (0.016)   & \B0.950 (0.018) & \B0.868 (0.030)   & 0.870 (0.057) & 3.60 \\
\bottomrule
\end{tabular}
\end{adjustbox}
\end{table}

\section{Conclusions and Future Work}

Our preliminary results highlight the room for improvement relative to other state-of-the-art LoRA-based federated learning methods. As we continue to develop our method we will explore the role of the rank of the global and local adapters, refine our approach to separating client-independent and client-specific knowledge between them, and expand our evaluations to larger models and additional tasks.

\subsubsection*{Acknowledgements}
This research was enabled in part by support provided by Calcul Qu\'ebec (\href{https://www.calculquebec.ca}{calculquebec.ca}) and the Digital Research Alliance of Canada (\href{https://alliancecan.ca/}{alliancecan.ca}). It was also supported through funding from Canadian Institute for Advanced Research (\href{https://cifar.ca/}{cifar.ca}) and the Natural Sciences and Engineering Research Council of Canada (\href{https://www.nserc-crsng.gc.ca/}{nserc-crsng.gc.ca}).

\bibliography{references.bib}
\bibliographystyle{iclr2025_conference}

\appendix

\section{Datasets}
\label{app:datasets}

\subsection{Fed-ISIC 2019}

The Fed-ISIC 2019 dataset \citep{ogierduterrail2022} contains images from four different clinics/hospitals in Australia, Austria, the USA and Spain. The images from the Austrian clinic are further divided by imaging machine for a total of six different clients. For our evaluations, we used the same test set as the FLamby benchmark \citep{ogierduterrail2022} and divided the training set for each client into training and validation sets using a stratified 80:20 split. The task is to classify images of skin lesions among nine different diagnostic categories: (0) melanoma, (1) melanocytic nevus, (2) basal cell carcinoma, (3) actinic keratosis, (4) benign keratosis, (5) dermatofibroma, (6) vascular lesion, (7) squamous cell carcinoma and (8) none of the others.

We pre-processed the images by resizing them to 224\;px by 224\;px (padding images where necessary to preserving the aspect ratio) and applying the same colour constancy normalization as the FLamby benchmark. We used random resized cropping, TrivialAugment \citep{muller2021} and random horizontal flipping for data augmentation during training.

\subsection{Camelyon17-WILDS}

The Camelyon17-WILDS is part of the WILDS domain generalization benchmark \citep{koh2021}. However, we use the dataset as a federated learning benchmark by simply treating each of the five hospitals as a separate client, as opposed to reserving some hospitals for training and other for testing. We divide the patients into training, validation and test sets using a 64:16:20 split. During splitting, we attempted to approximately balance the number of positive and negative examples across splits.

\begin{figure}[ht]
    \centering
    \includegraphics[width=0.7\linewidth]{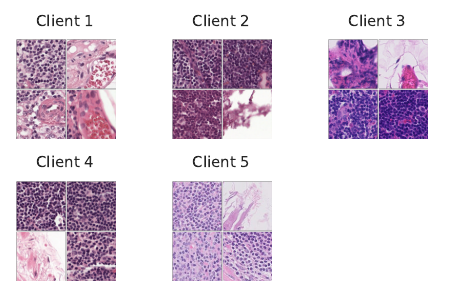}
    \caption{Patches of normal tissue from each client (hospital) in the Camelyon17 dataset showing the colour variation due to the differences in staining.}
    \label{fig:camelyon17-images}
\end{figure}

The dataset consists of 96\;px by 96\;px patches of whole-slide images with hematoxylin and eosin (H\&E) staining of a lymph node section from patients with possible metastatic breast cancer. Despite the images all using the same staining, slight differences in the stain and staining procedure between hospitals result in slight colour variations between clients. This can be seen in Fig. \ref{fig:camelyon17-images}.

The task is to classify the patches based on whether contain normal (0) or tumor (1) tissue. Tumor patches have at least one pixel of tumor tissue in the central 32\;px by 32\;px region and normal patches have no tumor and have at least 20\% normal tissue in the central 32\;px by 32\;px region. For data augmentation during training, we applied several augmentations that preserve the region of interest of the images. First, we took random 90\;px by 90\;px pixel crops which were resized to 224\;px by 224\;px. Then, we applied random translation (by up to 10\% in both the X and Y axes) and random rotation (by a multiple of 90 degrees). For testing, the patches were simply resized to 224\;px by 224\;px.

\section{Experimental Setup}
\label{app:experimental-setup}

For each task, we fine-tuned ViT-Tiny models from the PyTorch Image Models library \citep{wightman2019} that were pre-trained on the ImageNet-21K dataset. We used stochastic gradient descent (SGD) for all methods, with the learning rate $\eta \in \{1e^{-4}, 5e^{-4}, 1e^{-3}, 5e^{-3}, 1e^{-2}\}$ and momentum $\gamma \in \{0, 0.5\}$ tuned on the validation set. For the centralized and local baselines, the models were trained for 100 epochs and 20 epochs for the Fed-ISIC 2019 and Camelyon17-WILDS tasks, respectively. The federated learning methods were trained using FedAvg \citep{mcmahan2017} with 100\% client participation for 100 rounds and 20 rounds for the Fed-ISIC 2019 and Camelyon17-WILDS tasks, respectively. Each client performed one epoch of local training per round. Our experiments were performed using PyTorch and the Flower \citep{beutel2022} framework for the federated learning simulations.

The baselines against which we compared our methods differ in the way they use federated learning to train the LoRA adapters. FedIT \citep{zhang2024f} and FFA-LoRA \citep{sun2023c} do not perform personalization and train a single global adapter. FedIT trains both the $A$ and $B$ matrices of the LoRA adapter, whereas FFA-LoRA fixes the weights of the randomly initialized $A$ matrix and trains only the $B$ matrix. On the other hand, FedSA \citep{guo2024} and FedDPA \citep{yang2024c} are both personalized federated learning methods, like FedOPAL. FedSA trains both the $A$ and $B$ matrices of the LoRA adapter, but only the $A$ matrix is shared and updated by all clients. The $B$ matrix is personalized. Similar to FedOPAL, FedDPA trains separate global and a personal adapters.

\end{document}